\documentclass[10pt,twocolumn,letterpaper]{article}

\usepackage{wacv}
\usepackage{times}
\usepackage{epsfig}
\usepackage{graphicx}
\usepackage{amsmath}
\usepackage{amssymb}

\usepackage{booktabs}
\usepackage{listings}
\usepackage{caption}
\usepackage{subcaption}

\usepackage{listings}

\usepackage{xcolor}

\definecolor{darkergreen}{RGB}{21, 152, 56}
\definecolor{red2}{RGB}{252, 54, 65}
\newcommand\redp[1]{\textcolor{red2}{(#1)}}
\newcommand\greenp[1]{\textcolor{darkergreen}{(#1)}}

\definecolor{codegreen}{rgb}{0,0.6,0}
\definecolor{codegray}{rgb}{0.5,0.5,0.5}
\definecolor{codepurple}{rgb}{0.58,0,0.82}
\definecolor{backcolour}{rgb}{0.95,0.95,0.92}

\lstdefinestyle{mystyle}{
  backgroundcolor=\color{backcolour}, commentstyle=\color{codegreen},
  keywordstyle=\color{magenta},
  numberstyle=\tiny\color{codegray},
  stringstyle=\color{codepurple},
  basicstyle=\ttfamily\footnotesize,
  breakatwhitespace=false,         
  breaklines=true,                 
  captionpos=b,                    
  keepspaces=true,                 
  numbers=left,                    
  numbersep=5pt,                  
  showspaces=false,                
  showstringspaces=false,
  showtabs=false,                  
  tabsize=2
}

\wacvfinalcopy %

\ifwacvfinal
\def\assignedStartPage{1} %
\fi

\usepackage[pagebackref=true,breaklinks=true,colorlinks,bookmarks=false]{hyperref}

\ifwacvfinal
\else
\fi

\begin{document}

\title{StyleAugment: Learning Texture De-biased Representations \\by Style Augmentation without Pre-defined Textures}

\author{Sanghyuk Chun\\
NAVER AI Lab\\
\and
Song Park\\
Yonsei University\\
}

\maketitle

\begin{abstract}
Recent powerful vision classifiers are biased towards textures, while shape information is overlooked by the models. A simple attempt by augmenting training images using the artistic style transfer method, called Stylized ImageNet, can reduce the texture bias. However, Stylized ImageNet approach has two drawbacks in fidelity and diversity. First, the generated images show low image quality due to the significant semantic gap betweeen natural images and artistic paintings. Also, Stylized ImageNet training samples are pre-computed before training, resulting in showing the lack of diversity for each sample. We propose a StyleAugment by augmenting styles from the mini-batch. StyleAugment does not rely on the pre-defined style references, but generates augmented images on-the-fly by natural images in the mini-batch for the references. Hence, StyleAugment let the model observe abundant confounding cues for each image by on-the-fly the augmentation strategy, while the augmented images are more realistic than artistic style transferred images. We validate the effectiveness of StyleAugment in the ImageNet dataset with robustness benchmarks, such as texture de-biased accuracy, corruption robustness, natural adversarial samples, and occlusion robustness. StyleAugment shows better generalization performances than previous unsupervised de-biasing methods and state-of-the-art data augmentation methods in our experiments.
\end{abstract}

\section{Introduction}
While deep neural networks have shown the remarkable success comparing humans in complex vision recognition systems~\cite{resnet, ren2015faster}, deep neural networks have shown disappointed generalization performances against unfamiliar corruptions, such as noises, blurs, small perturbations, visual filters, or occlusions~\cite{imagenetc, geirhos2018generalisation, chun2019regeval}.
This fundamental limitation often weaken practical usages of deep models in real-world deployment scenarios, such as self-driving cars~\cite{michaelis2019benchmarking}.
As a naive approach for improving robustness against input distribution shifts, one can propose a data augmentation solution, \ie, augmenting corruptions during the training. However, the data augmentation approach is still not a perfect solution; a model trained with specific corruptions is only generalized to the seen corruptions, while the unseen corruption generalization is still not achievable~\cite{geirhos2018generalisation, chun2019regeval}.

Recent breakthroughs improving robustness have appeared at the different side of researches; a number of studies have shown that improving clean input performances also can help the robust representation against input corruptions~\cite{yun2019cutmix, xie2020adversarial, kim2020puzzle, yun2021relabel}. Especially, the most powerful models utilize a large amount of diverse extra data points~\cite{taori2020measuring} by semi-supervised learning with a very strong teacher~\cite{xie2020self, yun2021relabel} or learning with extra knowledge such as language supervision~\cite{clip}. However, learning with hundreds of millions of data points is not a always accessible solution for various visual recognition tasks, while learning without extra data still far from the performances of the clean images \eg, a model with 78.9\% clean accuracy only showing 28.1\% corrupted input accuracy~\cite{yun2021relabel}. This implies that we need more high-level understanding of why the deep vision models are not generalizable to unseen distribution shifts.

Recently, Geirhos \etal \cite{stylizedimagenet} have shown that strong vision classifiers, \eg, ResNet~\cite{resnet}, focus on the spurious texture cues, while shape information is overlooked by the network.
To mitigate the texture bias, Geirhos \etal \cite{stylizedimagenet} generated abundant texture-ized images by historical artistic paintings~\footnote{The authors used $\sim$80K artistic images from Kaggle's painter by numbers dataset (\url{https://www.kaggle.com/c/painter-by-numbers}).}, named Stylized ImageNet, based on artistic style transfer methods~\cite{neuralstyle, adain}.
By reducing the texture bias of neural networks using Stylized ImageNet, the shape-biased models show robust prediction on distribution shifts and better downstream transfer learning performances on object detection~\cite{ren2015faster}.
Bahng \etal \cite{bahng2019rebias} have shown that the existing deep models only focusing on small discriminative regions, resulting in being biased towards local cues, such as color and texture. By expanding the effective receptive field of the model, Bahng \etal \cite{bahng2019rebias} showed that the texture unbiased performances and performances under distribution shifts are improved.
From these observations, we presume that the texture bias is the source of unexpected behavior of deep neural networks.

\begin{figure*}
    \centering
    \includegraphics[width=0.22\linewidth]{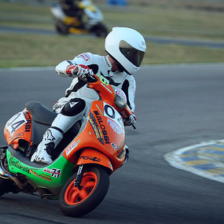}
    \includegraphics[width=0.22\linewidth]{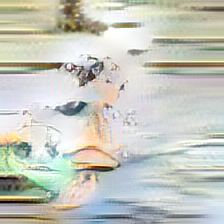}
    \includegraphics[width=0.22\linewidth]{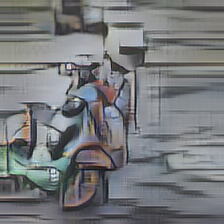}
    \includegraphics[width=0.22\linewidth]{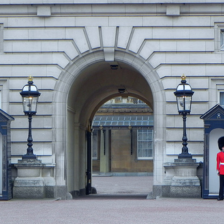}\\
    \includegraphics[width=0.22\linewidth]{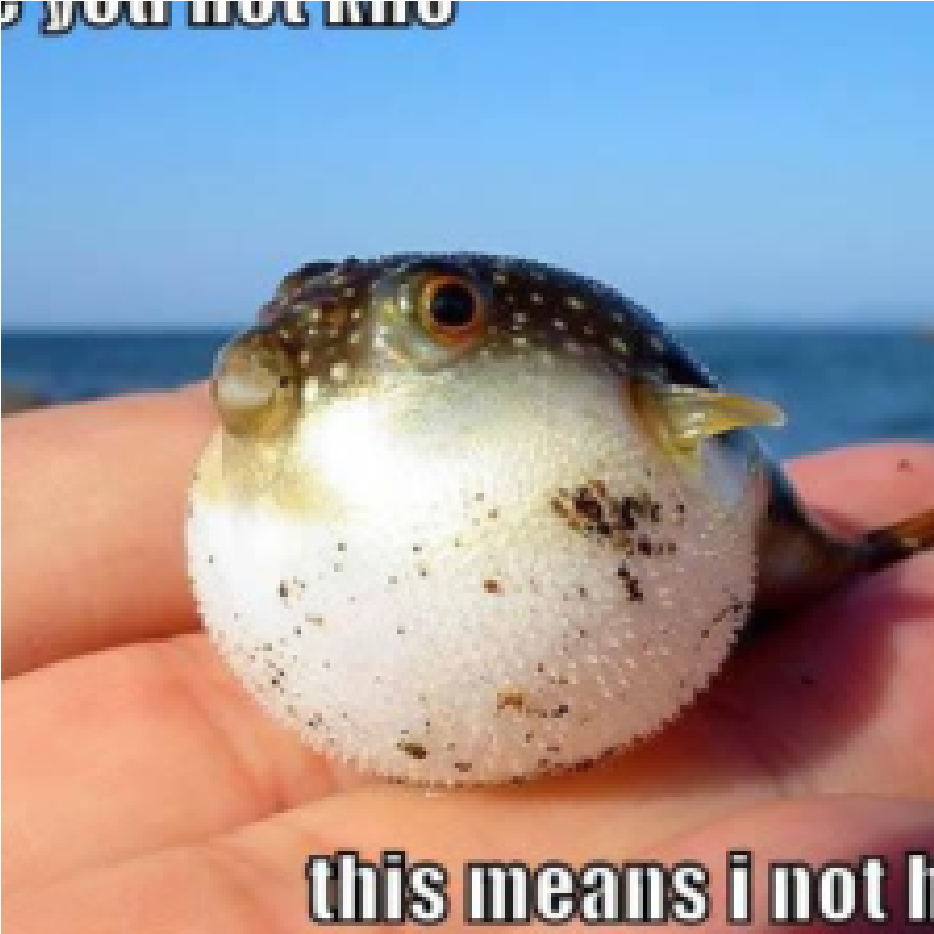}
    \includegraphics[width=0.22\linewidth]{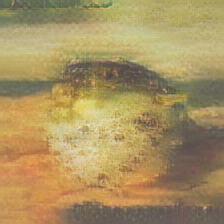}
    \includegraphics[width=0.22\linewidth]{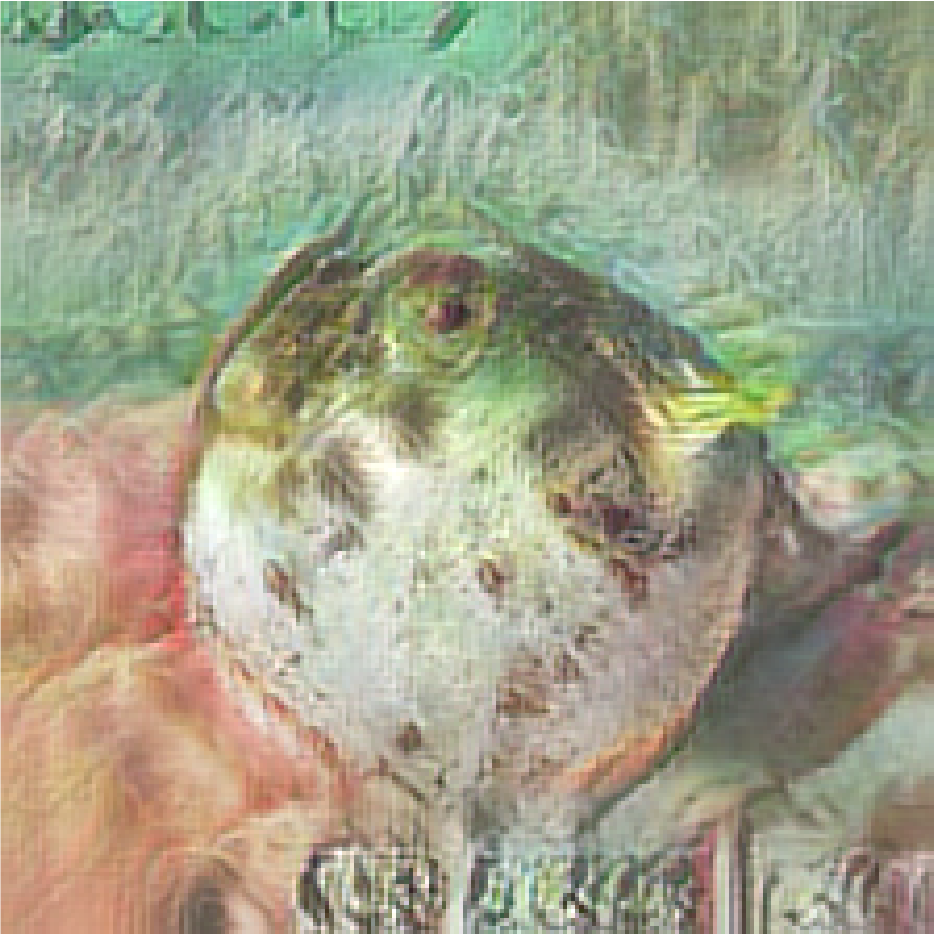}
    \includegraphics[width=0.22\linewidth]{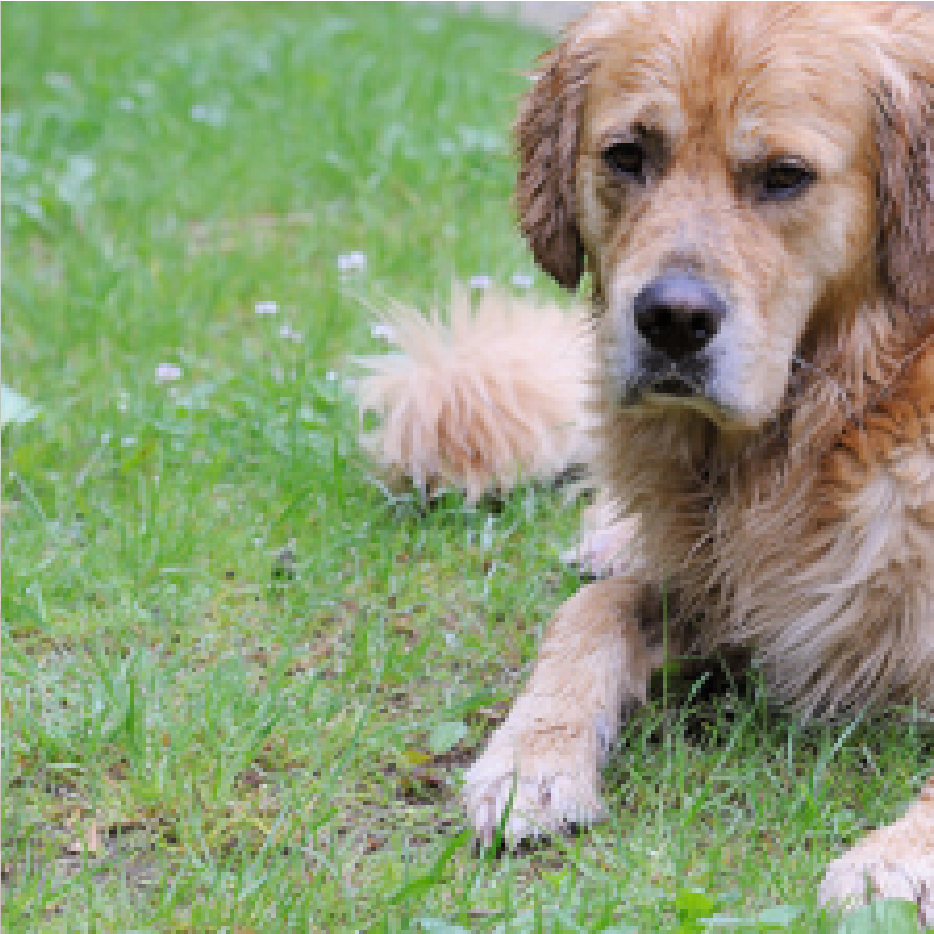}\\
    \includegraphics[width=0.22\linewidth]{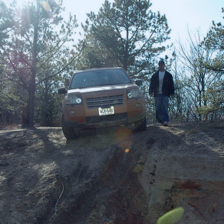}
    \includegraphics[width=0.22\linewidth]{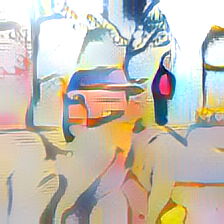}
    \includegraphics[width=0.22\linewidth]{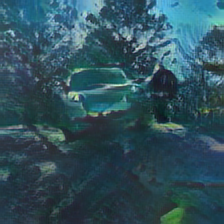}
    \includegraphics[width=0.22\linewidth]{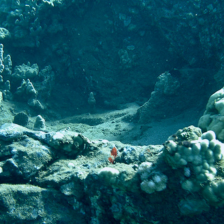}\\
    \includegraphics[width=0.22\linewidth]{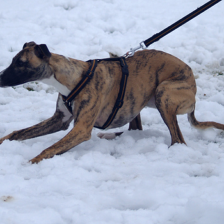}
    \includegraphics[width=0.22\linewidth]{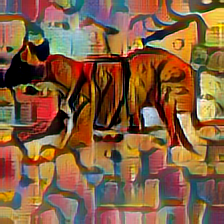}
    \includegraphics[width=0.22\linewidth]{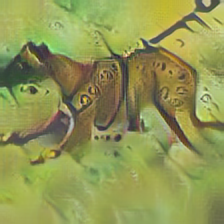}
    \includegraphics[width=0.22\linewidth]{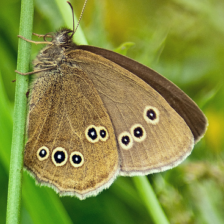}\\
    \begin{subfigure}[b]{0.22\linewidth}
        \includegraphics[width=\linewidth]{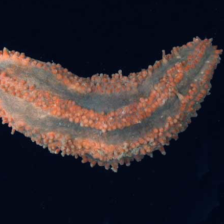}
        \caption{Clean Image}
        \label{fig:fig1a}
    \end{subfigure}
        \begin{subfigure}[b]{0.22\linewidth}
        \includegraphics[width=\linewidth]{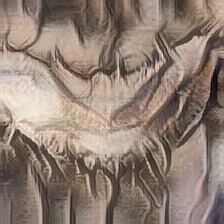}
        \caption{Artistic Stylization~\cite{stylizedimagenet}}
        \label{fig:fig1b}
    \end{subfigure}
        \begin{subfigure}[b]{0.22\linewidth}
        \includegraphics[width=\linewidth]{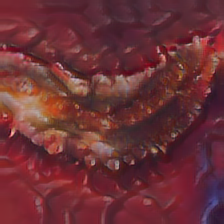}
        \caption{StyleAugment (Ours)}
        \label{fig:fig1c}
    \end{subfigure}
        \begin{subfigure}[b]{0.22\linewidth}
        \includegraphics[width=\linewidth]{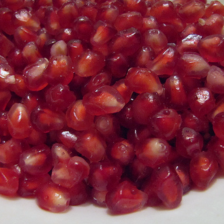}
        \caption{Style Reference Images}
        \label{fig:fig1d}
    \end{subfigure}
    \caption{\textbf{Limitation of Stylized ImageNet.} Due to the significant gap between the natural images (Figure \ref{fig:fig1a}) and the artistic images, the samples from Stylized ImageNet (Figure \ref{fig:fig1b}) (stylized by artistic images) often loses the content information, and shows low image fidelity. On the other hand, the samples generated by the proposed StyleAugmentation (Figure \ref{fig:fig1c}) utilizes natural images as the style reference (Figure \ref{fig:fig1d}) to achieve both preserving the content information and confounding the spurious correlations of the images, such as backgrounds.}
    \label{fig:teaser}
\end{figure*}

Previous attempts to mitigate texture bias is usually focusing on utilizing strong human inductive bias.
For example, Stylized ImageNet~\cite{stylizedimagenet} needs a pre-defined specification of texture images to augmentation.
Since it is unable to pre-define all possible textures in real-world, Geirhos \etal utilized a large historical artistic painting images for texture images.
However, as shown in Figure~\ref{fig:teaser}, because the significant gap between artistic images and natural images, the stylized images from artistic images often show low fidelity. As a result, solely training with Stylized ImageNet shows worse performances than training with clean ImageNet (76.13\% $\rightarrow$ 60.18\%~\cite{stylizedimagenet}). It shows that careful choice of texture images should be required for better performances.
Furthermore, due to the computation and the resource limitation, each image in Stylized ImageNet is only transferred by a random artistic image, \ie, not all $\sim$80K images in Kaggle's Painter by Numbers dataset are used for each image, but only one painting is chosen. This strategy can drastically reduce the diversity of the generated images. To sum up, the pre-computed stylized dataset has low fidelity due to the significant gap between natural images and artistic paintings and low diversity due to the dataset building strategy.

Similarly, Bahng \etal \cite{bahng2019rebias} also heavily relyed on the human inductive bias; the method should define an additional ``biased'' architecture by its design. For example, to reduce the texture bias of ResNet~\cite{resnet}, Bahng \etal \cite{bahng2019rebias} utilizes BagNet~\cite{bagnet} with restricted receptive field by changing the 3 $\times$ 3 convolution filters of ResNet to the 1 $\times$ 1 filters.
Although previous attempts with heavy human inductive bias have shown impressive improvements in many robustness benchmarks (\eg, ImageNet-A~\cite{imageneta}, ImageNet-C~\cite{imagenetc}, unbiased accuracies), it limits the practical usage in the tasks requiring a different inductive bias. Furthermore, theses methods can suffer from the carefully choice of pre-defined configurations, such as texture images~\cite{stylizedimagenet} or architectural choice~\cite{bahng2019rebias}, while a mis-specification can seriously drops overall performances.

In this paper, we propose a new data augmentation method, StyleAugment, for de-biasing texture biases of deep neural networks by augmenting styles from natural images. Unlike previous style augmentation method~\cite{stylizedimagenet}, our method does not require any pre-defined texture images to avoid heavy dependency on the quality of the pre-defined texture images and sensitivity to mis-specifications. Rather than augmenting styles from pre-defined images, our method extracts styles from the natural images from the training mini-batch.
Compared to the pre-stylization strategy by Geirhos \etal, our online stylization from the mini-batch strategy has two advantages; the style images are from the natural images as the content images which reduces the semantic gap between the transferred images and the original images; a model can observe diverse styles for each image while the pre-stylization strategy only allows to observe a specific pre-defined texture for each image.
StyleAugment also can be viewed as a Mixup variant \cite{mixup, yun2019cutmix}, but StyleAugment does not mix labels to let the model only focus on objectness, not confounding cues, such as backgrounds, textures, or color.

As a model trained with StyleAugment learns a texture de-biased representations, the model outperforms previous de-biasing methods (\eg, ReBias \cite{bahng2019rebias}, LfF \cite{nam2020lff}) and in-distribution data augmentation methods (\eg, CutMix \cite{yun2019cutmix}, Mixup \cite{mixup}) in ImageNet-9 \cite{bahng2019rebias} robustness benchmarks, such as shape-texture de-biased accuracy~\cite{bahng2019rebias}, corruptions (ImageNet-C~\cite{imagenetc}),  natural adversarial samples crawled from the web (ImageNet-A~\cite{imageneta}), and occluded samples.
Our experimental results on ImageNet-1k \cite{russakovsky2015imagenet} and CIFAR-10 \cite{cifar} show a similar conclusion; StyleAugment improves the robustness against the input distribution shifts, \eg, ImageNet-C or CIFAR-10-C benchmarks \cite{imagenetc}.
We also validate the design choice of StyleAugment in ImageNet-9. The results show that other design choices, such as changing the stylization method from AdaIN \cite{adain} to WCT$^2$ \cite{yoo2019wct2} and mixing labels as Mixup variants \cite{mixup, yun2019cutmix} improve the in-distribution generalization, but show performance drops in robustness benchmarks.

\section{Related Works}

\subsection{Unsupervised de-biasing}

While previous de-biasing method, such as fairness, assumes the existence of bias labels (\eg, assuming that there exists the protected attribute labels during the training), it is often unrealistic to assume the all bias labels are accessible by the model; labeling requires a huge human annotator costs, and the bias labels are often ill-defined, \eg, it is difficult to categorize natural textures in pre-defined texture sets.
To avoid the direct use of bias labels, de-biasing without explicit bias annotation has been actively studied in recent years. Especially, utilizing two networks, one for capturing bias and the other for de-biasing from the biased network, have been one of the major branches in this field. For example, in VQA tasks, the networks are known to be easily biased towards much easier text representations, while ignoring visual representations. To mitigate the issue, RUBi~\cite{cadene2019rubi} and LearnedMixin~\cite{clark2019learnedmixin} adjusted the final logit before the cross-entropy loss of the multi-modal model by using predictions from the uni-modal model where RUBi uses multiplication and LearnedMixinH uses summation. In the uni-modal debiasing tasks, HEX~\cite{hex} employs a texture extractor model and lets features extracted by the target model be orthogonal to the texture extractor outputs. ReBias~\cite{bahng2019rebias} makes two models have different capacity so a model is biased towards specific cues, \eg, color or texture. Learning from Failure (LfF)~\cite{nam2020lff} uses the re-weighting cross-entropy where the biased network is trained by the knowledge that networks in the early stage of training rely more on the spurious correlation than networks in the late stage of training.

While previous unsupervised de-biasing methods utilize additional biased model, our method does not need any additional biased model, but only requires a pre-trained style transfer algorithm, such as AdaIN~\cite{adain}. In the experiments, our method outperforms previous de-biasing methods in ImageNet clean accuracy as well as the ImageNet robustness benchmarks, such as texture de-biased accuarcy~\cite{bahng2019rebias}, ImageNet-C~\cite{imagenetc} and ImageNet-A~\cite{imageneta}.

\subsection{Data augmentation methods}

Since deep models are data hungry~\cite{InstagramNet}, synthesizing abundant training images by random crop, random flip, and color jittering have become a rule-of-thumb for enhancing the generalizability of deep models~\cite{InceptionResnet, densenet}.
Recent data augmentation methods have attended on either synthesizing mixed samples~\cite{mixup, yun2019cutmix}, or applying a series of very strong image filters, such as equalize, solarize, Cutout~\cite{cutout}, to make difficult samples~\cite{cubuk2019autoaugment, hendrycks2020augmix, cubuk2020randaugment}.
Our method can be viewed as a variant of Mixup approaches, such as Mixup~\cite{mixup}, CutMix~\cite{yun2019cutmix}, while our method does not mix labels of samples to prevent a model attending on spurious correlations, (\ie, textures) rather than the objectness (\ie, shape).
In this study, we did not compare our method with augmentation methods with visual filters, such as AugMix~\cite{hendrycks2020augmix} or RandAugment~\cite{cubuk2020randaugment}, because these algorithms need to pre-define augmentation types by a strong human prior knowledge, while our goal is to make universally applicable data augmentation algorithm without a strong inductive bias.

\subsection{Stylization-based augmentation methods}

Since Geirhos \etal \cite{stylizedimagenet} have shown that deep convolutional neural networks are biased towards textures, and a simple stylization-based augmentation (StyleImageNet) can mitigate the texture biases, stylization-based augmentation methods have been studied in a robustness view point.
While the previously proposed data augmentation methods showed that the performance improvements in existing robustness benchmarks~\cite{mixup, yun2019cutmix, hendrycks2020augmix}, the data augmentation methods for in-distribution generalization is known to be not helpful when the semantic gap between the training images and the test images is significantly large, such as domain generalization benchmarks~\cite{cha2021swad}.
From this motivation, Zhou \etal \cite{zhou2021mixstyle} proposed a Mixup-like stylization augmentation method named MixStyle for domain generalization benchmark. MixStyle mixes two images from different domains to let the model be generalized to diverse domains. Our method focuses on a conventional image recognition task (\ie, the ImageNet classification task) in the de-biasing aspects (\ie, additional ImageNet robustness benchmarks). The stylization-based augmentation strategy also studied by Hong \etal \cite{hong2021stylemix}, a contemporary work of our study. Hong \etal proposed to utilize both the content loss and the style loss by mixing labels as Mixup~\cite{mixup} or CutMix~\cite{yun2019cutmix}. Our method does not mix the labels to avoid confounding factors, \ie, textures.

Stylized ImageNet \cite{stylizedimagenet} is proposed to mitigate the texture bias of deep models, not only learning shape-biased features but also showing performance improvements in downstream tasks, such as ImageNet classification, object detection, and the ImageNet-C corruption robustness benchmark.
Stylized ImageNet, however, has fundamental limitations on its image fidelity (as the significant gap between the artistic paintings and the target natural images) and lack of the style diversity for each image (as the data generation process generating all training images before training). Our method mitigates these two issues by utilizing the natural images from the mini-batch as the style reference, showing better fidelity and diversity as shown in Figure~\ref{fig:teaser}.

\section{Style Augmentation Without the Pre-defined Styles or Textures}

\subsection{Preliminary: Arbitrary Style Transfer by Adaptive Instance Normalization (AdaIN)}

Arbitrary style transfer tasks~\cite{neuralstyle, adain, wct} aim to generate an image with the given two images, \ie, content and style images. The underlying assumption of style transfer methods since Gatys \etal \cite{neuralstyle} is that the feature statistics, including mean and standard deviation, represent the texture of images. Primal studies~\cite{neuralstyle} directly optimize the input image to match the feature statistics of the content (mean) and the style (covariance, or Gram matrix) in an iterative manner.
Real-time style transfer methods, such as whitening-coloring-transform (WCT) \cite{wct} or adaptive instance normalization (AdaIN) \cite{adain}, approximate the optimization process by replacing feature statistics of content and style images. We use AdaIN style transfer method for real-time data augmentation.

Let $Enc$ and $Dec$ be an encoder and a decoder. AdaIN style transfer first extracts a feature $f$ from the input image $x$ by $f = Enc(x)$. For the content and style images $c, s$ and their corresponding features $f_c, f_s$, the AdaIN operation is defined as follows:
\begin{equation}
\label{eq:adain}
    \text{AdaIN}(f_c, f_s) = \sigma(f_s) \left( \frac{f_c - \mu(f_c)}{\sigma(f_c)} + \mu(f_c) \right),
\end{equation}
where $\mu(\cdot), \sigma(\cdot)$ denotes the feature statistics from the instance normalization.
The transferred feature is decoded by the decoder as $\tilde x = Dec(AdaIN(f_c, f_s))$.
We use the ImageNet-trained VGG-16 network~\cite{vgg} as the encoder $Enc$, and the official AdaIN decoder as the decoder $Dec$ following the official implmentation\footnote{\url{https://github.com/xunhuang1995/AdaIN-style}}.

\begin{figure*}
\begin{lstlisting}[language=Python]
def train_iteration(inputs, targets):
    # inputs: a standard Torch image array
    # targets: a standard Torch label array
    rand_index = torch.randperm(inputs.size()[0])
    
    # styletransfer: an arbitrary function for style transfer. The former argument is the content and the latter is the style.
    transferred = styletransfer(inputs, inputs[rand_index])
    inputs = torch.cat([inputs, transferred], dim=0)
    targets = torch.cat([targets, targets], dim=0)

    # model is a regular CNN and criterion is the cross entropy function
    outputs = model(inputs)
    loss = criterion(outputs, targets)
    
    # optim: a standard optimizer such as Adam or AdamP
    optim.zero_grad()
    loss.backward()
    optim.step()
\end{lstlisting}
\caption{PyTorch pseudo code for StyleAugment}
\label{fig:algorithm}
\end{figure*}
\subsection{StyleAugment: a stylization-based augmentation without pre-defined textures}

For the given mini-batch $\mathcal B = (x_1, x_2, \ldots, x_n)$ with batch size $n$, StyleAugment generates a new augmented mini-batch $\mathcal B'$ by augmenting stylized images where the style references are from the other samples in the mini-batch.
In each training iteration, StyleAugment randomly combines styles from the mini-batch to make diverse image samples.
In practice, we train the model with the original mini-batch and the augmented mini-batch, \ie, $\mathcal B + \mathcal B'$.
We illustrate the PyTorch pseudo code for StyleAugment in Figure~\ref{fig:algorithm}.
StyleAugment does not require any prior knowledge on the pre-defined textures (\eg, Stylized ImageNet~\cite{stylizedimagenet}) or strong image visual filters (\eg, AutoAugment~\cite{cubuk2019autoaugment}, AugMix~\cite{hendrycks2020augmix}, RandAugment~\cite{cubuk2020randaugment}).
StyleAugment training strategy enables the mini-batch-level knowledge interaction as Mixup variants~\cite{mixup, yun2019cutmix}, but StyleAugmentation does not mix labels which can make the model rely on spurious correlations, such as textures or backgrounds.
We will discuss the details in \S\ref{sec:abl}.

\subsection{Discussions}
\label{subsec:limit}

Compared to the artistic stylized images, natural stylized images show benefits on the image fidelity. Figure~\ref{fig:teaser} shows the example stylized images by artistic style transfer (Figure~\ref{fig:fig1b}) and by the proposed StyleAugment (Figure~\ref{fig:fig1c}). Interestingly, the images generated by StyleAugment show diverse distribution shifts, rarely observed in real-world. For example, in the second row of the figure, the background of the globefish seems as a grassfield. Similarly, in the fourth row, the dog shows the ``eye'' texture from the butterfly image, and the background changes from snow to the grass-like texture.
In other words, StyleAugment generates an image with uncommon correlations, such as ``a globefish on the grassfield'' or ``a dog with butterfly patterns''.
By augmenting rare combinations of the true objectness and the other confounding factors, StyleAugment makes a model learn de-biased representations to the spurious correlations, such as background, textures, or color.

However, StyleAugment still has a limitation on its image quality. For example, as shown in the first row of Figure~\ref{fig:teaser}, StyleAugment tends to preserve the original content information compared to artistic style transfer, but there exists the damage of the content information. This may hurt the final performance of the model as shown in \cite{stylizedimagenet}. To understand the trade-off between shape-preserving and stylization, we evaluate our method with a photorealistic style transfer method~\cite{yoo2019wct2}, focusing on the edge preserving, but only transferring color information. Our experimental results show that StyleAugment with AdaIN shows worse results in in-distribution generalization performances, but better results in distribution shift generalization. We will discuss the details in \S\ref{sec:abl}.

\section{Experiments}

In this section, we demonstrate the effectiveness of StyleAugment in ImageNet classification tasks. We also compare our design choice with other possible variants of StyleAugment.

\subsection{Experiments Settings}

\begin{table*}[ht]
\centering
\begin{tabular}{@{}lccccc@{}}
\toprule
Model               & Clean & Unbiased Acc~\cite{bahng2019rebias} & ImageNet-C~\cite{imagenetc} & ImageNet-A~\cite{imageneta} & Occlusion \\ \midrule
Vanilla (ResNet-18~\cite{resnet})$^\dagger$ & 90.8  & 88.8        & 54.2       & 24.9       & 71.3      \\
Biased (BagNet-18~\cite{bagnet})$^\dagger$  & 67.7  & 65.9        & 31.7       & 18.8       & 59.7      \\ \midrule
LearnedMixin + H~\cite{clark2019learnedmixin}$^\dagger$        & 64.1  & 62.7        & 27.5       & 15.0       & 33.5      \\
RUBi~\cite{cadene2019rubi}$^\dagger$                & 90.5  & 88.6        & 53.7       & 27.7       & 71.3      \\
ReBias~\cite{bahng2019rebias}$^\dagger$              & 91.9  & 90.5        & 57.5       & 29.6       & 73.4      \\
LfF~\cite{nam2020lff}                 & 93.2	& 92.0	& 57.8	& 28.1	& 77.0         \\ \midrule
CutMix~\cite{yun2019cutmix}              & \underline{93.8}  & 91.8        & 54.6       & 27.1       & \textbf{83.1}      \\
Mixup~\cite{mixup}               & 93.2  & 91.4        & 61.5       & \textbf{33.4}       & \underline{77.9}      \\
Stylized ImageNet~\cite{stylizedimagenet}$^\dagger$   & 88.4  & 86.6        & 61.1       & 24.6       & 64.4      \\ \midrule
StyleAugment            & \underline{93.8}  & \underline{92.6}        & \underline{65.3}       & 29.6       & 73.0      \\
StyleAugment + AdamP~\cite{heo2021adamp}    & \textbf{95.9}	& \textbf{94.8}	& \textbf{72.5}	& \underline{32.1}	& 75.8\\ \bottomrule
\end{tabular}
\caption{\textbf{Comparison of state-of-the-art de-biasing and augmentation methods on the ImageNet-9 validation dataset}. We measure the ImageNet-9 top-1 validation accuracy (Clean), the unbiase accuracy using texture clustering (Unbiased Acc) following Bahng \etal \cite{bahng2019rebias}, ImageNet-C top-1 accuracy, ImageNet-A top-1 accuracy, and the top-1 accuracy on occluded samples. The first and the second best methods are denoted in \textbf{bold numbers} and \underline{underlined numbers}. The rows with $^\dagger$ denotes the same weights from Bahng \etal \cite{bahng2019rebias}.}
\label{tab:main}
\end{table*}

\begin{table*}[ht]
\centering
\begin{tabular}{@{}lccc|cccc@{}}
\toprule
                 & Clean & ImageNet-A~\cite{imageneta} & ImageNet-C~\cite{imagenetc} & Noise & Blur & Weather & Digital \\ \midrule
ResNet-18        & 69.8  & 1.1        & 30.1             & 30.8  & 18.9 & 12.3    & 9.4     \\
\, + StyleAugment & 68.3 \redp{-1.5}  & 2.1 \greenp{+1.0} & 35.8 \greenp{+5.6}             & 39.0 \greenp{8.1} & 25.1 \greenp{6.2} & 19.9 \greenp{7.6}    & 17.2 \greenp{7.8}    \\ \midrule
ResNet-50        & 76.1  & 0.8        & 36.2             & 41.0  & 26.1 & 19.3    & 16.5    \\
\, + StyleAugment & 73.8 \redp{-2.3}  & 3.5 \greenp{2.7}       & 43.6 \greenp{7.4}             & 53.2 \greenp{12.1}  & 38.7 \greenp{12.6} & 34.5 \greenp{15.1}    & 32.8 \greenp{16.3}    \\ \bottomrule
\end{tabular}
\caption{\textbf{Impact of StyleAugment on the ImageNet-1k validation dataset.} Clean accuracy, ImageNet-A top-1 accuracy, ImageNet-C top-1 accuracy, and the average performances on the ImageNet-C subsets are shown. We use the official ResNet models provided by the PyTorch vision library. Note that rows with ``+StyleAugment'' are trained without conventional image distortions such as color jittering and lightening, while the baseline methods are trained with the image distortions.}
\label{tab:imagenet}
\end{table*}

\paragraph{Dataset.}
We use the ImageNet \cite{russakovsky2015imagenet} classification benchmarks for measuring the effectiveness of our method. In the main experiments, we use the subset of ImageNet with 9 super-class (ImageNet-9) as Bahng \etal \cite{bahng2019rebias}. ImageNet-9 consists 54,600 training images and 2,100 test images.
We also measure the generalization performance of the models using distribution shifted ImageNet images.
First, we measure the \textbf{unbiased accuracies} of ImageNet-9 as Bahng \etal. The unbiased accuracy measures the average of combination-wise accuracies where the combination is found by the texture clustering algorithm. The unbiased accuracy is computed by taking an average over all possible combinations of texture clusters and image labels. Showing better unbiased accuracy means the model is less biased towards spurious texture information.
\textbf{ImageNet-C}~\cite{imagenetc} contains 20 corruptions\footnote{The original ImageNet-C paper suggests to use 15 corruptions for evaluation, while 5 corruptions are remained as ``test'' set. We use all 20 corruptions to compute ImageNet-C performances of the models.}, such as Gaussian noise, motion blur, weather changes. We measure the average performances over 20 corruptions and 5 severties. An improved performance on ImageNet-C implies that 
\textbf{ImageNet-A}~\cite{imageneta} is a collection of the failure cases of the ImageNet-trained ResNet-50 \cite{resnet} model (called ``natural adversarial examples'').
As previous studies \cite{taori2020measuring, xie2020self, yun2021relabel, clip} have observed, achieving high ImageNet-A performances without extra dataset is a challenging task, without considering architectural changes \cite{heo2021pit}. Therefore, better ImageNet-A accuracy (without extra dataset or architecture changes) indirectly shows that the model less relies on shortcuts in the datasets for their predictions.
Finally, we report the performances of the \textbf{center occluded} images for measuring occlusion robustness. Following \cite{yun2019cutmix}, we zero-ed out the center pixels with the 112 $\times$ 112 patch size.

\paragraph{Implementation details:}
For fair comparisons, we follow the setting of Bahng \etal \cite{bahng2019rebias} for ImageNet-9 experiments; We use the ResNet-18 \cite{resnet} backbone with the batch size as 128, the initial learning rate as 0.001, and the cosine learning rate scheduling. The models are trained with 120 epochs. We exclude all image distortion-based augmentations, such as color jittering and lightening for precisely understanding the effectiveness of each method.
For ImageNet-1k experiments, we use the same setting as ImageNet-9, and the models are trained with 90 training epochs. We trained ResNet-18 and ResNet-50 with StyleAugment for ImageNet-1k.

Finally, we additionally report CIFAR-10 \cite{cifar} results, where the batch size is set to 128, the number of training epochs is set to 100, and the initial learning rate is set to 0.1 decayed by the cosine annealing scheduling.

Note that StyleAugment doubles the number of training samples as shown in Figure~\ref{fig:algorithm}. We set the number of clean samples as the half of the batch size in all experiments.

All experiments except ImageNet-9 adopts AdamP optimizer \cite{heo2021adamp} (ImageNet-9 experiments use Adam \cite{kingma2015adam}). We use 2 V100 GPUs for ImageNet experiments, and 1 V100 GPU for CIFAR-10 experiments.
NAVER Smart Machine Learning (NSML) \cite{nsml} is used for all experiments.

\paragraph{Comparison methods.}
We compare our StyleAugment with two major directions of researches: unsupervised de-biasing methods and data augmentation without human prior knowledge.
We compare StyleAugment with four unsupervised de-biasing methods, including LearnedMixin + H \cite{clark2019learnedmixin}, RUBi \cite{cadene2019rubi}, ReBias \cite{bahng2019rebias}, and Learning from Failure (LfF) \cite{nam2020lff}. These methods are designed to overcome the shortcut learning (\ie, bias problems) of the models.
We also evaluate three data augmentation methods in the same benchmark: CutMix \cite{yun2019cutmix}, Mixup \cite{mixup} and Stylized ImageNet \cite{stylizedimagenet}. Other augmentation methods requiring extra knowledge on image distortions (\eg, solorize, equalize) are not compared in this study to avoid unexpected effects by the additional image distortions.

\begin{table}[]
\centering
\begin{tabular}{@{}lcc@{}}
\toprule
      & CIFAR-10 test & CIFAR-10-C \\ \midrule
ResNet-18 & 92.3         & 66.1      \\
\, +StyleAugment      & 92.3 \greenp{+0.0}     & 67.6 \greenp{+1.5}     \\ \bottomrule
\end{tabular}
\caption{\textbf{Impact of StyleAgument on CIFAR-10 dataset.} CIFAR-10 test accuracy and CIFAR-10 corrupted (CIFAR-10-C) performances are shown.}
\label{tab:cifar}
\end{table}
\subsection{Main results}

\paragraph{ImageNet-9.}
Table \ref{tab:main} shows the comparison of de-biasing methods, data augmentation methods, and our StyleAugment.
In the table, we first observe that data augmentation methods show remarkable improvements in in-distribution accuracies (93.8\% by CutMix and 93.2\% by Mixup) comparing to de-biasing methods, such as ReBias and LfF, but their unbiased accuracies are worse than LfF (92.0\% by LfF, 91.8\% by CutMix and 91.4\% by Mixup).

In ImageNet-C benchmarks, de-biasing methods and CutMix shows marginal improvements against the baseline (baseline: 54.2\%, ReBias: 57.5\%, LfF: 57.8\%, CutMix: 54.6\%), while Mixup and Stylized ImageNet show significant ImageNet-C performance improvements (Mixup: 61.5\%, Stylized ImageNet 61.1\%). In particular, Stylized ImageNet shows worse performances in clean, unbiased accuracy, ImageNet-A and occlusion benchmarks, but shows remarkable performance improvement in ImageNet-C. This implies that stylization helps the robustness against common corruptions, but its low visual quality hurts the in-distribution generalization performance as well as other unbiased performances.

We also observe that ReBias (24.9\% $\rightarrow$ 29.6\%) and LfF (24.9\% $\rightarrow$ 28.1\%) show remarkable improvements in ImageNet-A performances, while CutMix (27.1\%) and Stylized ImageNet (24.6\%) show marginal improvements compared to ReBias and LfF. This result implies that de-biasing methods are helpful for improving robustness against spurious correlations.

Finally, we observe that our StyleAugment shows the best performances in in-distribution generalization (same as CutMix -- 93.8\%), unbiased accuracy (92.6\%, while 91.8\% CutMix is the second best one), and ImageNet-C benchmark (with a large margin to 61.5\% by Mixup, StyleAugment shows 65.3\%). Although StyleAugment shows worse ImageNet-A performances, StyleAugment shows the same ImageNet-A performances with ReBias. Since StyleAugment focuses on learning feature distribution shifts, StyleAugment only shows marginal improvements in occlusion benchmark (71.3\% $\rightarrow$ 73.0\%).

We also report the results trained by StyleAugment and AdamP optimizer \cite{heo2021adamp} showing strong performance improvements in various tasks, including computer vision, robustness, and audio tasks. By using AdamP, we improve the performances of StyleAugment by significant gaps for clean accuracy (93.8\% $\rightarrow$ 95.9\%), unbiased accuray (92.6\% $\rightarrow$ 94.8\%), ImageNet-C performance (65.3\% $\rightarrow$ 72.5\%), ImageNet-A performance (29.6\% $\rightarrow$ 32.1\%), and occlusion benchmark (73.0\% $\rightarrow$ 75.8\%).

\begin{table*}[ht]
\centering
\begin{tabular}{@{}lccccc@{}}
\toprule
Model               & Clean & Unbiased Acc~\cite{bahng2019rebias} & ImageNet-C~\cite{imagenetc} & ImageNet-A~\cite{imageneta} & Occlusion \\ \midrule
StyleAugment (proposed) & 93.8 & 92.6 & 65.3 & 29.6 & 73.0 \\ \midrule
StyleAugment + WCT$^2$\cite{yoo2019wct2} & 94.1 \greenp{+0.3} & 92.2 \redp{-0.4} & 57.0 \redp{-8.3} & 31.7 \greenp{+2.1} & 77.6 \greenp{+4.6} \\
StyleAugment + Label mixing & 94.2 \greenp{+0.4}	& 92.7 \greenp{+0.1}	& 62.0 \redp{-3.3}	& 29.3 \redp{-0.3}	& 76.0 \greenp{3.0} \\ \bottomrule
\end{tabular}
\caption{\textbf{Ablation study on design choices of StyleAugment.} We compare the peformances of StyleAugment variants on ImageNet-9 benchmark as Table~\ref{tab:main}.}
\label{tab:abl}
\end{table*}

\paragraph{ImageNet-1k and CIFAR-10}
Table \ref{tab:imagenet} shows the impact of StyleAugment in the ImageNet-1k benchmark. Note that our StyleAugment results did not use the standard augmentations, such as color jittering and lightening. In the table, we observe that the StyleAugmented models show slightly worse performances in the in-distribution generalization performances (due to the lack of the standard augmentations). However, in other robustness benchmarks including ImageNet-A and ImageNet-C, the StyleAugmented models show better performances than the vanilla counterparts even with worse clean accuracies.
We observe similar results in Table \ref{tab:cifar} for CIFAR-10 experiments. The StyleAugmented model shows comparable clean accuracy, but better corruption robustness compared to the vanilla ResNet-18.

\paragraph{Implication and limitation.}
As we observed in ImageNet-9 and CIFAR-10-C experiments, our StyleAugment improves the overall performances of the deep models when the number of training data points is small (54K for ImageNet-9, 50K for CIFAR-10). On the other hand, ImageNet-1k results show that StyleAugment improves the robustness of the model, while the model shows performance drops in clean accuracy.
We presume that it is because our StyleAugment implementation does not contain the standard color jittering and lightening augmentations.
Also, we presume that it is due to the low image fidelity of AdaIN transferred images. As shown in Geirhos \etal \cite{stylizedimagenet}, the performance can be improved by applying additional ``fine-tuning'' process on the clean training images. We did not test fine-tuning strategy in this study, but we assume the image quality affects a lot to the in-distribution performances. As a primal study, in the following section, we investigate the effect of the style transfer method to in-distribution generalization and out-of-distribution generalization.

\subsection{Ablation studies}
\label{sec:abl}

We conduct ablation studies of design choices of StyleAugment in ImageNet-9. We tested two variants of StyleAugment in the image quality and the target label.
First, as we discussed in the previous sections, the image quality by AdaIN is not perfect (Figure \ref{fig:teaser}). Especially, AdaIN loses edge and shape information of the original image. We use a photorealistic style transfer model WCT$^2$, focusing on preserving edge information by the Haar wavelet transform.
Note that WCT$^2$ transfers color and light information, while AdaIN transfers texture information.
In the second row of Table \ref{tab:abl}, we report the StyleAugment performances when AdaIN encoder and decoder modules are changed to WCT$^2$ encoder and decoder. For real-time computation, we modify the original WCT$^2$ algorithm from whitening-coloring-transform \cite{wct} to adaptive instance normalization (Equation \eqref{eq:adain}).
We observe that StyleAugment with WCT$^2$ shows better in-distribution generalization performance (94.1\%) than StyleAugment with AdaIN (93.8\%), as well as ImageNet-A (29.6\% $\rightarrow$ 31.7\%) and occlusion performances (73.0\% $\rightarrow$ 77.6\%). However, StyleAugment with WCT$^2$ shows drastic performance drops in ImageNet-C (65.3\% $\rightarrow$ 57.0\%) by losing texture information.
To sum up, the better image quality by content preservation improves in-distribution generalization performances, but cannot generalize the corruption robustness by losing advantages of texture transferring.

We also test a variant on the target labels of StyleAugment. First, we mix content and style labels for the target label as mixup variants. We observe similar phenomenon to the results of WCT$^2$; it improves clean accuracy, but drops ImageNet-C performance. We assume that it is because StyleAugment allows the model to observe a sample with various spurious cues, such as texture and background (as shown in Figure \ref{fig:teaser}). However, if we mix the content and style labels, the model can attend unexpected prediction cues, rather than object information itself.

\section{Conclusion}
In this paper, we propose a new data augmentation method using stylization methods. The proposed StyleAugment generates augmented images by applying AdaIN style transfer between the mini-batch samples, while the previous stylization-based approach, Stylized ImageNet, uses pre-defined artistic paintings. Compared to Stylized ImageNet, the model trained with our StyleAugment can observe more diverse and realistic images with various confounding factors such as backgrounds, textures.
In the experiments, we show that our StyleAugment shows improvements in robustness benchmarks, such as corruption robustness, while showing comparing or outperforming in-distribution generalization performances.
While changes in the stylization method or the label mixing strategy improve the in-distribution generalization performances, the changes show worse robustness performances. It shows that our StyleAugment strategy makes images with various spurious correlations from style images, \eg, texture, resulting in improvements of robustness performances.

{\small
\bibliographystyle{ieee_fullname}
\bibliography{egbib}
}

\end{document}